\documentclass[letterpaper, 10 pt, conference]{ieeeconf}  

\IEEEoverridecommandlockouts                              

\usepackage{pdfpages}
\usepackage{float}
\usepackage{subcaption}
\usepackage{hyperref}

\usepackage{graphics} 
\usepackage{graphicx} 
\title{\LARGE \bf
TELESIM: A Modular and Plug-and-Play Framework for Robotic Arm Teleoperation using a Digital Twin
}

\author{Florent P Audonnet$^{1*}$, Jonathan Grizou$^{2*}$, Andrew Hamilton$^{3}$ and Gerardo Aragon-Camarasa$^{4*}$
\thanks{This research has been supported by EPSRC DTA No. 2605103 and EPSRC Grant No. EP/S019472/1}
\thanks{$^{*}$School of Computing Science, University of Glasgow, G12 8QQ, Scotland, United Kingdom}
\thanks{$^{1}${\tt\small f.audonnet.1@research.gla.ac.uk}}%
\thanks{$^{2}${\tt\small jonathan.grizou@glasgow.ac.uk}}
\thanks{$^{3}${\tt\small andrew.w.hamilton@glasgow.ac.uk}}
\thanks{$^{4}${\tt\small gerardo.aragoncamarasa@glasgow.ac.uk}}%
}

\begin{document}
\let\oldUrl\url
\renewcommand{\url}[1]{\href{#1}{Link}}

\maketitle
\thispagestyle{empty}
\pagestyle{empty}

\begin{abstract}

We present TELESIM, a modular and plug-and-play framework for direct teleoperation of a robotic arm using a digital twin as the interface between the user and the robotic system. We tested TELESIM by performing a user survey with 37 participants on two different robots using two different control modalities: a virtual reality controller and a finger mapping hardware controller using different grasping systems. Users were asked to teleoperate the robot to pick and place 3 cubes in a tower and to repeat this task as many times as possible in 10 minutes, with only 5 minutes of training beforehand. Our experimental results show that most users were able to succeed by building at least a tower of 3 cubes regardless of the control modality or robot used, demonstrating the user-friendliness of TELESIM.

\end{abstract}

\section{Introduction}\label{sec:intro}

Robot teleoperation is difficult for non-experts~\cite{rea_still_2022, muto_touch_2012}. Recently, the ANA Avatar XPRIZE Challenge~\cite{noauthor_xprize_2023} set a series of challenging tasks to test the limits of teleoperation. The best systems that completed the challenges were rewarded with a prize pool of \$10 million. At its core, the challenge involves the direct teleoperation of a robot with minimal latency and the capacity to experience the environment from the robot's perspective. 
However, direct teleoperation still places a heavy physical and mental strain on the user, as Pettinger \textit{et al.}  \cite{pettinger_reducing_2020} reported that a user performing a pick and place task was faster and had fewer errors while reporting the task was more accessible when shared autonomy systems were enabled. Hence, researchers from HCI, medicine, robotics, and others, have explored different means of control for teleoperation to address these limitations. While most research efforts focus on a physical control device such as a Virtual Reality Controller \cite{dafarra_icub3_2022, rakita_motion_2017}, a Joystick \cite{scherzinger_learning_2023, javdani_shared_2015}, or phone \cite{mandlekar_human---loop_2020}, others decided to use cameras to track the whole body \cite{chen_intuitive_2012, lin_shared_2020}, or just the gaze \cite{admon_predicting_2016}. There has yet to be an overall consensus on the most appropriate type of control for direct teleoperation with specific applications requiring specific implementations.

In this paper, we develop a modular and plug-and-play direct teleoperation framework called TELESIM that non-experts can use without specialised training using off-the-shelf Virtual Reality (VR) technologies. Specifically, TELESIM objective is to allow for the direct teleoperation of any robotic arm using a digital twin as the interface between the user and the robotic system. We then demonstrate TELESIM's user-friendliness using a user study and the users' success rate at completing the task using two different types of control and grasping systems. Specifically, we use a virtual reality controller and a finger mapping hardware controller mounted on two robotic manipulators using different grasping systems. We compare their performance to study whether additional degrees of freedom in the control scheme enhance performance while performing a simple task. Our contributions are:

\begin{itemize}
    \item A modular and plug-and-play framework for teleoperation for any robotic arm using a digital twin.
    \item An experimental validation for testing the framework's performance through a simple non-expert task.
    \item A rigorous evaluation involving 37 participants demonstrating the user-friendliness of TELESIM.
\end{itemize}

\begin{figure}[t]
    \centering
    \includegraphics[width=0.85\linewidth]{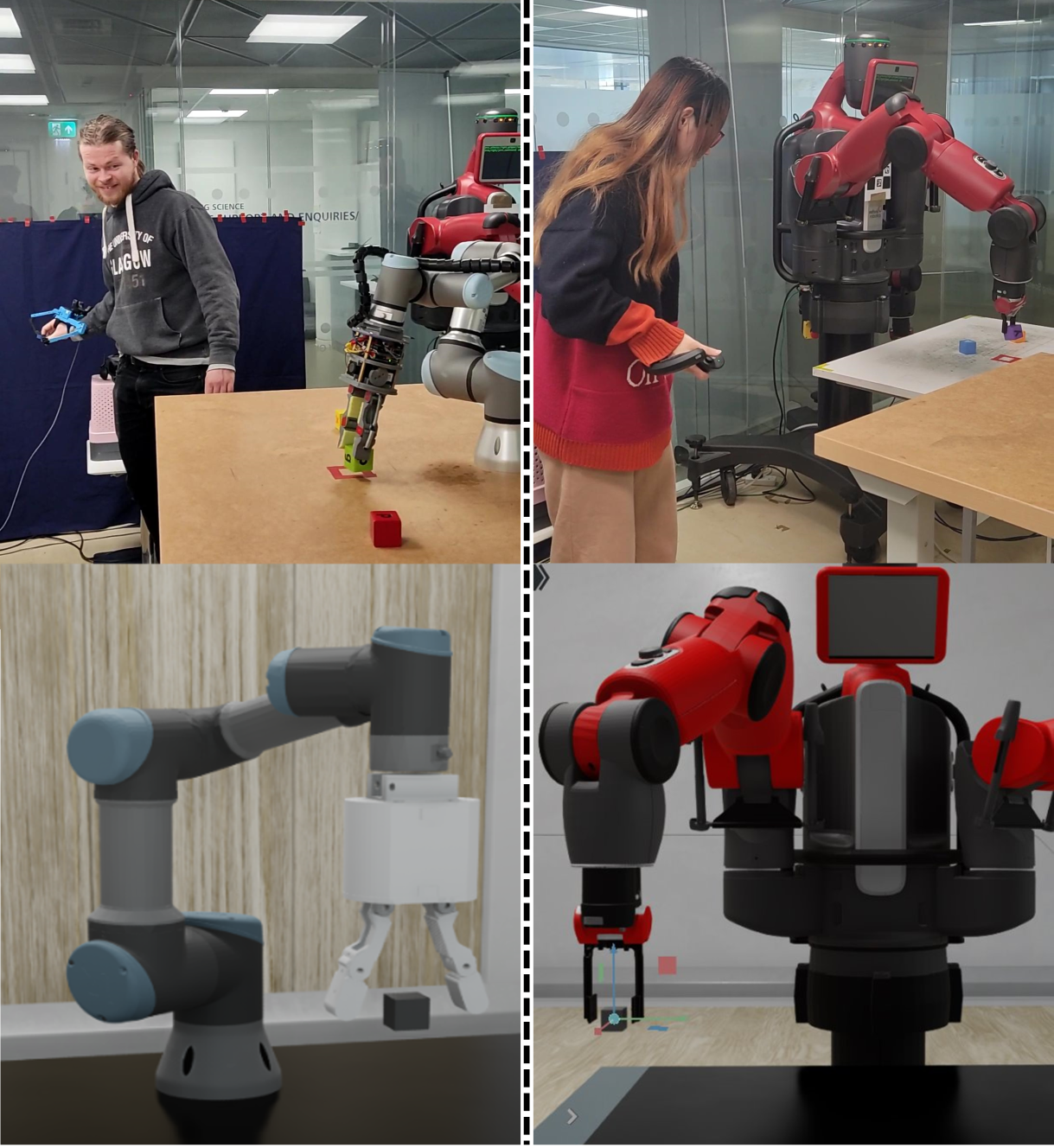}
    \caption{Our modular and plug-and-play TELESIM framework is being used to control a UR3 Robot (top-left) and a Baxter Robot (top-right) and its digital twin (bottom-right). The robot's digital twins can be seen underneath their respective real robots}
    \label{fig:isaac_sim}
\end{figure}

\begin{figure*}[t]
\centering
\includegraphics[width=0.9\textwidth]{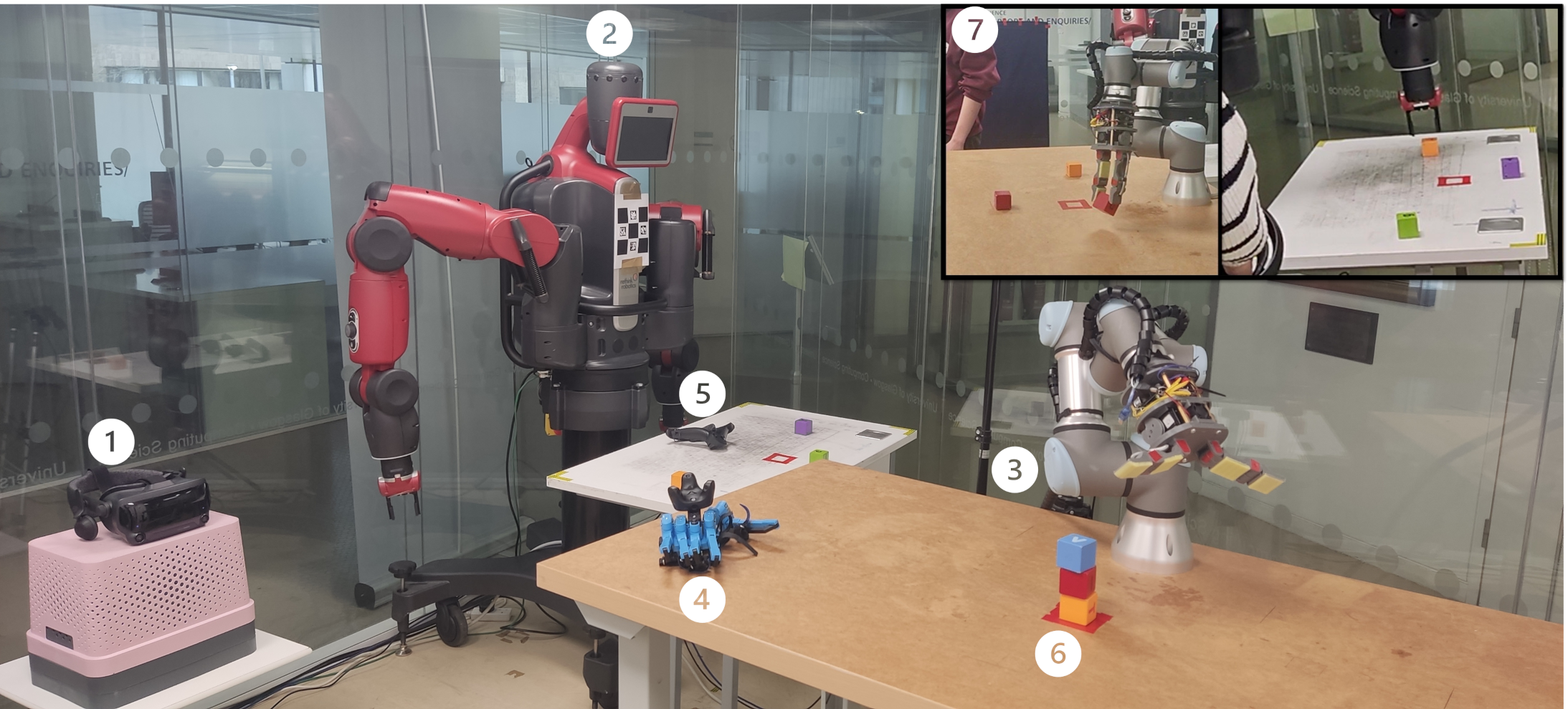}
\caption{Overview of the experimental setup. The Steam Index VR Headset \cite{noauthor_valve_nodate} is marked as (1) on the far left, which acts as the world's origin. The Baxter robot on the left (2) is controlled by the Steam Index controller (5). In front of it, the UR3 is on the right (3), with the Yale OpenHand T42 gripper \cite{noauthor_yale_2023}, controlled by the Senseglove and HTC Vive tracker (4) on the left side of the brown table. Additionally, in the upper right corner (7), a bird eye view of the task, which consists of 3 cubes in a triangle pattern (described in Section \ref{sec:experimental_setup}), while on the brown table, the cubes are arranged in the goal configuration (6)}
\label{fig:overview}
\end{figure*}

\section{Background}

Direct teleoperation is considered a stepping stone for shared autonomy \cite{zhang_haptic_2021}. This is because direct teleoperation causes significant cognitive strain on the user \cite{pettinger_reducing_2020}, and the user may not be capable of millimetre-scale adjustment to the position of the robot end effector. While in medicine, the user's movement is scaled down to allow for more precision \cite{lanfranco_robotic_2004, rakita_motion_2017}, it may not be suitable for all types of manipulation tasks as some require significant arm movements to move an object from one place to another. 

Hence, researchers have explored different control methods to reduce the cognitive strain while giving the highest amount of precision. For instance, low degree-of-freedom control methods such as a keyboard \cite{katyal_approaches_2014}, a joystick \cite{aronson_eye-hand_2018, scherzinger_learning_2023}, a touchscreen \cite{toh_dexterous_2012}, or a gamepad\cite{micire_design_2011} have brought an improved level of control \cite{katyal_approaches_2014} to address the user's mental strain. However, with the advent of VR technologies, researchers have investigated whether these technologies are appropriate for direct teleoperation. For example, they have proposed using a VR controller such as \cite{ wang_intent_2021, dafarra_icub3_2022} or a phone \cite{mandlekar_roboturk_2018}. While others have investigated the use of motion mapping of the user's body \cite{chen_intuitive_2012, rosen_communicating_2019} or only gaze control \cite{admon_predicting_2016}. However, for the latter, the added mobility generates a higher cognitive load \cite{pettinger_reducing_2020}, and mapping motions to robot movements is challenging due to differences in kinematics chains between robot arms and users \cite{kennel-maushart_manipulability_2021}.

Recently, Gottardi \textit{et al.} \cite{gottardi_shared_2022} have investigated combining multiple control systems, such as a VR controller and a tracking band on the upper arm, to track the user's movements. Rakita \textit{et al.}\cite{rakita_motion_2017} also compared various control methods; a stylus, a touchscreen, and a VR controller. These were then integrated into a custom inverse kinematics solver that adjusted the tolerance level when matching the end-effector pose to that of the user. The authors showed that users preferred the VR controller as they were more successful at completing pick-and-place tasks, such as picking up bottles or plates.

To mitigate the limitation of direct teleoperation, researchers have focused on how much shared autonomy improved the success of a given task. For this, research works have aimed at comparing direct teleoperation with respect to an assisted version to analyze the impact of shared autonomy on task success. For example, Chen \textit{et al.} \cite{chen_intuitive_2012} created a system in which the operator, using a joystick, manipulated the robot's end effector to an object, and then the robot could either grasp the object autonomously or assist the user in fine-tuning the robot position for a more optimal grasp. Later,
\cite{lin_shared_2020, pettinger_reducing_2020, gottardi_shared_2022} built on \cite{chen_intuitive_2012} where the user teleoperated the robots directly to a planned position but allowed the robot to perform the grasp automatically or, in the case of \cite{pettinger_reducing_2020}, turn a valve handle. Lin \textit{et al.} and Gottardi \textit{et al.} conducted a user survey and confirmed that users preferred the shared autonomy approach, as it reduced complexity and mental strain. Furthermore, \cite{gottardi_shared_2022} observed results similar to \cite{javdani_shared_2015}, who hypothesised that users preferred to give up control if it meant increasing the task completion rate. However, Javdani \textit{et al.}~\cite{javdani_shared_2015, javdani_shared_2018} have falsified this hypothesis using a system similar to \cite{lin_shared_2020, pettinger_reducing_2020, gottardi_shared_2022}. The authors concluded that users preferred to lose control if it meant an increase in a task's success rate only for a more complex task, while, for simple tasks, users still preferred to have more control. These works have focused on one robotic system and conducted their experimental survey on a small user base (between 8 and 12 participants). Furthermore, they focused on different autonomy levels and not on different control methods.

\begin{figure*}[!t]
\centering
\includegraphics[width=0.9\textwidth]{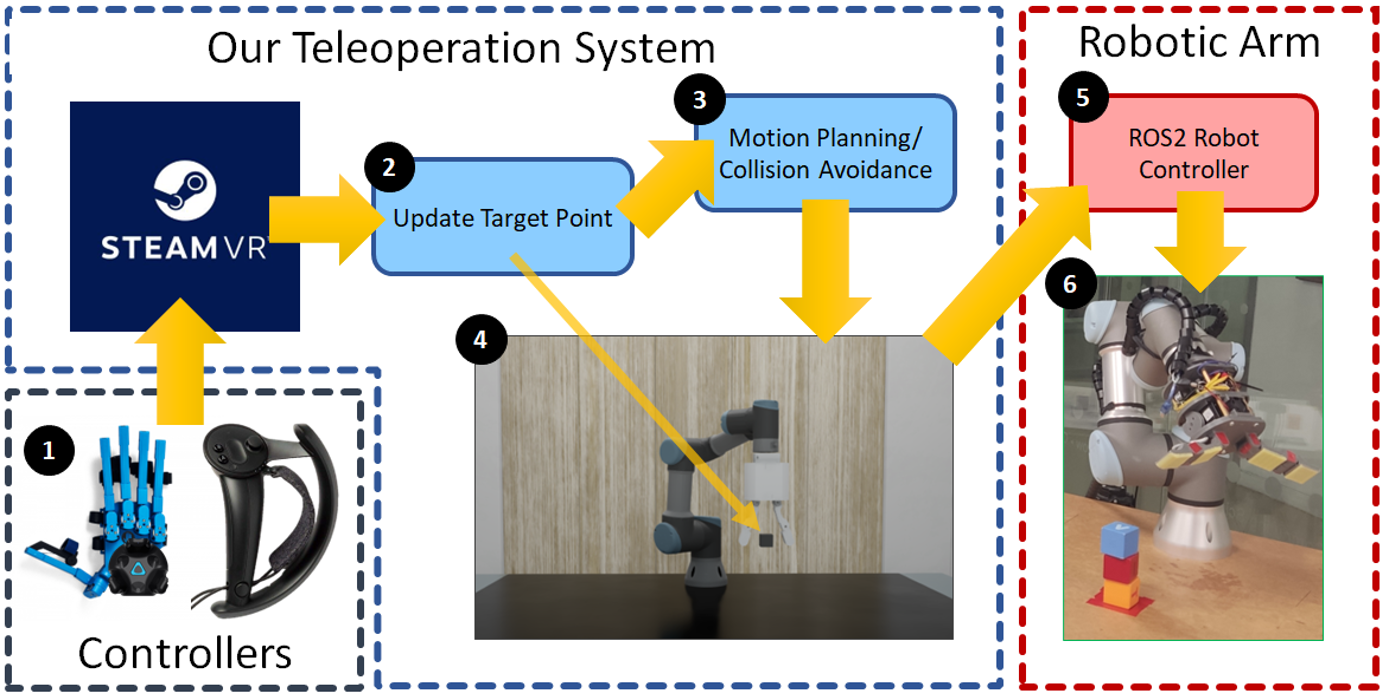}
\caption{Overview of TELESIM, our direct teleoperation framework. The controllers (in the black dotted line) can be any system that outputs a 3D pose, while our framework is depicted in the blue dotted line. Finally, TELESIM can be plugged into any robotic system via a ROS2 robot controller, as shown in the red dotted line.}
\vspace{-0.5cm}
\label{fig:schema}
\end{figure*}

Although this paper focuses on TELESIM as a framework, our evaluation also addresses two main limitations of previous work: (1) researchers have only used one robot per study, and (2) most user studies consider a small user base, which does not represent a statistically significant sample. It also addresses a gap discussed by Rea \& Seo \cite{rea_still_2022}, which states that there needs to be more non-expert evaluation of robotic teleoperation for general tasks such as picking and placing common objects. Therefore, to advance the state-of-the-art in robotic teleoperation, we investigate the performance of different control modalities for direct teleoperation using a VR controller and finger mapping. Similarly, we evaluate our framework on 2 different robots, a Rethink Robotics Baxter and a Universal Robotic 3, and ask 37 non-expert participants to carry out a simple pick-and-place task. Additionally, by using two different control modalities, our goal is to bridge the gap between VR control methods\cite{ wang_intent_2021, dafarra_icub3_2022} and complete body mapping \cite{chen_intuitive_2012,  lin_shared_2020} by investigating the performance of hand and finger tracking through a SenseGlove.

\section{Direct Teleoperation Framework}\label{sec:framework}

TELESIM (Fig. \ref{fig:schema}) consists of 3 components. The controller (1) can be any type of control system that outputs a 3D pose. In this paper, the controller is either a VR controller or a SenseGlove with an HTC Vive Tracker. Our framework (2) calculates a motion plan to the pose given by (1) and sends it through ROS2 (Robot Operating System) \cite{macenski_robot_2022} to any type of robotic arm that supports ROS control (3). In this paper, we use either a Baxter robot or a UR3 robot. The following subsections detail the design and implementation of each component.

\subsection{Hardware Setup}\label{sec:hard_setup}

The hardware used in this paper consists of a SteamVR~\cite{noauthor_steamvr_nodate} tracking a Steam Index controller and HTC Vive Tracker, as shown in Figs. \ref{fig:overview} points 5 and 4, respectively. The Baxter robot (point 2 in Fig. \ref{fig:overview}) is controlled using the Steam Index controller, and the gripper can be closed by pressing the main trigger on the Steam Index controller. The UR robot (point 3 in Fig. \ref{fig:overview}) is controlled using a Senseglove development kit, which allows for mapping of all individual finger movements, with an HTC Vive Tracker mounted on top and is linked to a T42 gripper from the Yale OpenHand project \cite{noauthor_yale_2023}. These control methods are represented as point 1 in Fig. \ref{fig:schema}

The T42 gripper is an underactuated 2-finger gripper that we modified to have a 1 degree of freedom per finger. The interface with the Senseglove is thus restricted to movements of the user's thumb and index finger, and these are mapped to a single degree of freedom to result in the opening and closing of the gripper. We adopted this approach to compare the Steam Index controller and the Senseglove fairly, but still allowed us to test if the increase in complexity of adding an extra degree of freedom on the user side has effects on the success rate.

We also developed a custom controller for the gripper using two separate Arduino UNOs, as they only have one serial port per board. Joint positions are converted from degrees to motor commands, which are then sent to the first Arduino through a serial interface. Then, commands are relayed to the second Arduino through I2C and transmitted to the two MX-28 Dynamixels\cite{noauthor_robotis_nodate}. The second Arduino is also responsible for reading the motors' current position and load and transmitting it back to the computer. 
Additionally, we restricted the amount of force the user can transmit to the gripper, as without it, it leads to breakage of the controlling string or exceeding the amount of resistance allowed by the motor. Since we were interested in developing a simple task that non-experts can carry out, we decided not to implement haptic feedback as it would give the users an advantage of sensing whether an object is grasped. Thus, this will result in an unfair comparison between the VR controller and the SenseGlove. Therefore, we leave haptic feedback for future work.

The VR headset (1 in Fig. \ref{fig:overview}) acts as the origin of both robots, giving the user an easy reference point for teleoperation. The SteamVR outputs the controller position in a 3D space with respect to the headset. The origin is thus transformed into the user's resting hand position when initiated. This method of tracking the position is preferred by multiple researchers\cite{pettinger_reducing_2020, whitney_ros_2018}, as well as many of the participants in the ANA Avatar XPRIZE Challenge \cite{dafarra_icub3_2022, schwarz_robust_2023}. The Senseglove is also used by the winning team \cite{schwarz_robust_2023}, but to control a Schunk robotic hand that replicates a human hand.

\subsection{Digital Twin}\label{sec:digital-twin}

The position in the 3D space from the SteamVR is transmitted through ROS2 to a full digital twin created in NVIDIA Isaac Sim\cite{noauthor_isaac_2019}. This flow of information can be seen in Fig. \ref{fig:schema} point 2. Isaac Sim, a recent ray-tracing simulation software, is used to calculate the robot motion plan using RMPFlow~\cite{cheng_rmpflow_2020}, which is a motion generation based on Riemannian Motion Policies~\cite{ratliff_riemannian_2018} (Fig. \ref{fig:schema} point 3). Isaac Sim was chosen as it is the most realistic simulation software compatible with ROS2 from our simulation benchmark~\cite{audonnet_systematic_2022}.

Isaac Sim takes in a URDF of a robot for visualisation along with a robot description file, describing the joints that can be actuated by the motion planner and the robot's collision as spheres, as Isaac Sim uses them for collision checking. This file can be created with an included extension (Lula Robot Description Editor\cite{noauthor_lula_nodate}). The ability to add new robots is a significant part of what makes TELESIM modular, along with ROS2.

We decided to use ROS2 as it is the most used framework for controlling a variety of robots, making our framework plug-and-play. Views of the UR3 and the Baxter robot from Isaac Sim are shown in \ref{fig:isaac_sim}. The grey square in between the gripper (Fig. \ref{fig:schema} point 4) is the point that is controlled by the teleoperation system and indicates where Isaac Sim must find a path. 

The fact that Isaac Sim acts as a complete digital twin allows the robot to avoid collision with the world around it and damage itself. Additionally, both robots have systems that allow them to work alongside humans. Our system is capable, with minimal configuration, of handling the restrictions placed by their need to be safe around humans. Finally, Isaac Sim transmits the position of each joint to the real robot through ROS2 and passes this information to the robotic system, as shown in Fig. \ref{fig:schema} points 5 and 6.

\subsection{Robot Control}

ROS control \cite{chitta_ros_control_2017}, provides a wrapper that facilitates the interface of different robots. However, each robot needs to be adapted to work with specific hardware. For this paper, we implemented the Universal Robot ROS2 Control package and used it to transfer the joint states from Isaac Sim to the robots (Fig. \ref{fig:schema} point 5). 

Specifically, for the UR3, the size of the gripper and the safety regulations of the laboratory reduced the available workspace for the robot. We handled these limitations by adding safety planes and limitations on the range of motion of the real robot. For Baxter, we used a ROS1 to ROS2 bidirectional bridge, as Baxter only works with ROS1 internally. The pipeline is thus converting Isaac Sim's joint states into Baxter messages in ROS2 and then they are sent through the bridge as ROS1 messages to the robot. The robot outputs its current state, which is converted to ROS2 through the same bridge.

Our framework introduces a slight amount of lag (500 ms) between the user movement and the robot movement, partly due to the path planning step, the time taken by the actuator to move the robot arm to the desired position and mostly due to safety restrictions that caps the maximum speed of our robots. The faster the user moves from one position to another, the higher the delay as the arm tries to catch up. This is easily accounted for by the user by making small and slow movements, as confirmed by Schwarz and Behnkle \cite{schwarz_low-latency_2021}. Although we acknowledge lag is present in our system, Schwarz and Behnkle \cite{schwarz_low-latency_2021} found that minor delays do not impact performance, and none of our users mentioned lag as an issue in our experiments (see Section \ref{sec:eval}).


\section{Experiments}\label{sec:experimental_setup}

\subsection{Methodology}

As described in Section \ref{sec:hard_setup}, the user controls a robot by teleoperating it using either a VR controller or the SenseGlove. For this experiment, we consider a simple task where the user has to pick up 3 cubes on a table. The user then needs to bring them individually to the centre of the table to complete a tower. Once the tower is completed, the cubes are returned to their original position, and the user is asked to repeat the task as many times as possible within 10 minutes.

The task definition consists of 3 cubes positioned at each vertex of an isosceles triangle and at similar distances from the robots' base on each robot's table (Fig. \ref{fig:schema} point 7). We have placed markers of the vertices of this isosceles triangle on each robot's operating table for repeatability of the task between attempts and among users. The front cube at the top of the triangle is positioned such that it is at the maximum reach limit of each robot's overhead rotation. This means that the end-effector's z-axis is perpendicular to the table, which makes it difficult for the user to pick up the cube from overhead. Thus, the user has to add some rotation in the x- or y-axis of the gripper to pick the cube successfully.
The right cube is the furthest away from the user in both robots and adds a degree of difficulty due to the user's viewpoint, but still within reach. Finally, the left cube is placed such that the user has to move their body. That is, for Baxter, the location where the user needs to pick up the left cube is approximately at the waist of the user, while for the UR3, the location is on the other side of (farthest from) the headset. The cube positions were chosen to let users be spatially aware of their position and its relationship to the robot. Finally, the location where the users need to stack the three cubes is easily accessible by the robot, and this position is marked by a red square in red tape, 2 cm larger than the cube. This position can be seen in Fig. \ref{fig:overview} point 6, with the cubes stacked as required. 

In order to teleoperate using the Baxter robot, the users need to stand with their back to the VR headset, while for teleoperating using the UR3, users need to stand with the headset on their right and the Baxter robot behind them. This difference in position is due to the space constraint of the room in which we ran our experiments. The operating room can be seen in Fig. \ref{fig:overview}, with the headset on the left of the picture shown as point 1.

\subsection{User Survey}
In our experiments, we asked 37 participants (29 male and 8 female) from various backgrounds aged 19 to 51 (mean: 25.32, 1 standard deviation: 6.26) to teleoperate both robots and stack 3 cubes without a monetary reward. Participants reported having, on a 5-point Likert scale (going from "Experienced" with a score of 1 to "No Experience" with a score of 5), a 3.03 mean experience with Virtual Reality with a standard deviation of 1.2. They also reported having a mean experience of 3.24 with a robot with a standard deviation of 1.24

Each participant completed the short questionnaire described above at the beginning of the experiment. After being asked to position their back to the VR headset, an explanation was given on how to control the robot, emphasising that all of their hand movements and rotation will be mapped one-to-one to the robot. They were instructed to try to grasp a cube from both sides. They had 5 minutes to get used to the control without a specific task objective. Most of the participants picked up and placed a cube during this time. 

After 5 minutes, the participants performed the task of stacking the 3 cubes in the given location without any restriction on the cubes' pose and order. Users were asked to stack cubes as many times as possible in 10 minutes. Once a tower has been completed, we reset the cubes to their initial configuration. Users' actions were recorded, such as the time taken for each tower and for individual actions for each pick, place, and drop (i.e. failures). 

After 10 minutes, users were given the option to take a break while answering the Single-Ease Question (SEQ) \cite{hodrien_review_2021}. Then, they were asked to repeat the same experiment but with the UR3 robot. SEQ was chosen instead of other metrics such as the System Usability Scale (SUS)~\cite{brooke_sus_1996} as Hodrien and Fernando~\cite{hodrien_review_2021} have argued that it is a good end-of-task metric.

\section{Evaluation}\label{sec:eval}

\begin{figure}[t]
\includegraphics[width=0.95\linewidth]{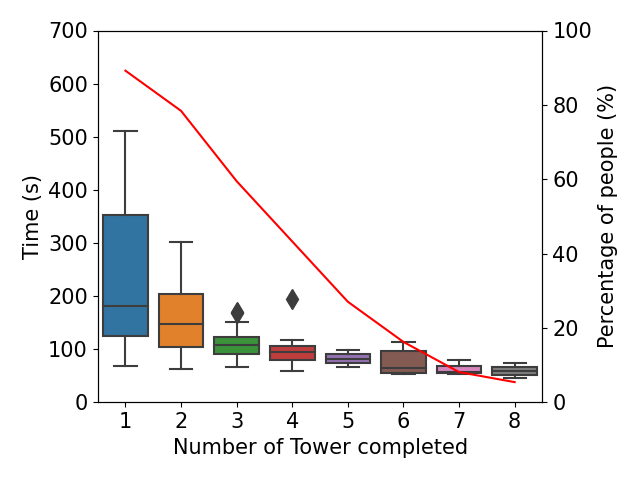}
\caption{Average Time Taken and Percentage of Population for each Tower Completed for Baxter}
\label{fig:baxter_tower}
\end{figure}

\begin{figure}[t]
\includegraphics[width=0.95\linewidth]{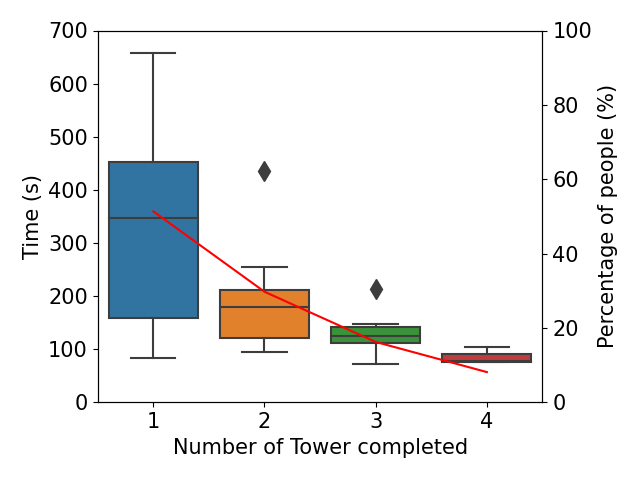}
\caption{Average Time Taken and Percentage of Population for each Tower Completed for UR}
\label{fig:ur_tower}
\end{figure}

Fig. \ref{fig:baxter_tower} shows that $85\%$ of the participants can build at least one tower in 10 minutes using Baxter and the VR controller. However, there is a steady decline for each of the following towers, with only $5\%$ of the users able to build 8 towers. This is in direct comparison to the UR3 as shown in Fig. \ref{fig:ur_tower}, with slightly less than $50\%$ of the population failing to build one tower and $5\%$ managing to build 4, half as many towers for Baxter.

The box plots in Fig. \ref{fig:baxter_tower} and \ref{fig:ur_tower} show the average and variance of the time taken by users to complete the towers. In particular, the first tower for both robots took most of the task duration because some participants could not build one tower. 
This time completion trend shows TELESIM's user-friendliness as $60\%$ of users for Baxter managed with minimal training to complete a full tower, which means 3 different pick-and-place operations in around 2 minutes. Similarly, this can also be observed for the UR3 as $25\%$ of the users managed to build a tower in 4 minutes. 

\begin{table}[t]
     \vspace{1mm}
     \caption{Additional Statistics Collected}
     \label{tab:stats}

    \begin{subtable}[h]{0.45\textwidth}
        \caption{Baxter }

        \centering
        \renewcommand{\arraystretch}{1.5}
        \begin{tabular}{|c|c|c|c|}
        \hline
    
     & Min & Mean $\pm$ Std & Max \\
    \hline
    Placing Rate & 25.00\% & 77.42\% $\pm$ 15.54\%  & 100.00 \\
    Dropping Rate & 3.70\% & 23.83\% $\pm$ 14.08\% & 66.67 \\
    Collapse Rate & 5.56\% & 18.44\% $\pm$ 11.66\% & 57.14 \\
    Still in Place Rate & 24.31\% & 75.21\% $\pm$ 15.20\% & 95.92 \\
    \hline
        \end{tabular}
        \label{tab:baxter}
        \end{subtable}
    \hfill
    \begin{subtable}[h]{0.45\textwidth}
            \hspace{0.5pt}
            \caption{UR}

        \centering
        \renewcommand{\arraystretch}{1.5}
        \begin{tabular}{|c|c|c|c|}
        \hline
    
     & Min & Mean $\pm$ Std & Max \\
    \hline
    Placing Rate & 12.50\% & 46.29\% $\pm$ 17.97\%  & 86.67 \\
    Dropping Rate & 13.33\% & 53.37\% $\pm$ 18.63\% & 87.50 \\
    Collapse Rate & 4.76\% & 22.25\% $\pm$ 12.46\% & 50.00 \\
    Still in Place Rate & 14.88\% & 46.93\% $\pm$ 16.75\% & 84.44 \\
    \hline
        \end{tabular}
        \label{tab:ur}
         \end{subtable}
\end{table}


Table \ref{tab:stats} shows the additional statistics collected during the experiment, such as the percentage of times the user dropped a cube that caused the tower to collapse. Table \ref{tab:baxter} indicates that for the Baxter robot, $75\%$ of the picking actions resulted in a correctly placed cube that did not collapse due to incorrect placement or the user inadvertently moved the robot in the tower's path. Similarly, in Table \ref{tab:ur}, $46\%$ of all the picking actions resulted in a correct place. The difference in the number of towers built, shown in Fig. \ref{fig:ur_tower}, can be explained by a greater difficulty in picking the cube. Specifically, our results indicate that there is no significant difference in the difficulty of picking a cube ($P>0.05$), nor is there a difference in the amount of time that the user collapses a tower while placing a cube ($P>0.1$). This lack of difference shows the stability of the teleoperation, as the difference in placing rates and the number of towers can be explained by the difference in control modality and the difference in robots. 

The collapse rate in both Table \ref{tab:baxter} and Table \ref{tab:ur} is similar and indicates that the type of robot does not influence the difficulty in safely placing a cube in a specific spot. However, the difference in drop rate for the UR3 can be related to the limitations of the gripper described in Section \ref{sec:hard_setup}, such as the limitation of the grip strength of the closed finger, to prevent the cable from breaking, and the limited range of motions, since we observed that some users let the cube fall while the gripper was closed. However, successful users moved slowly to prevent unnecessary movement, thus reducing the risk of dropping. This limitation is also visible in the placing rate; the placing and dropping rates are complementary, as these are the only two outcomes after picking up a cube.

\begin{figure}[tb]
\includegraphics[width=0.45\textwidth]{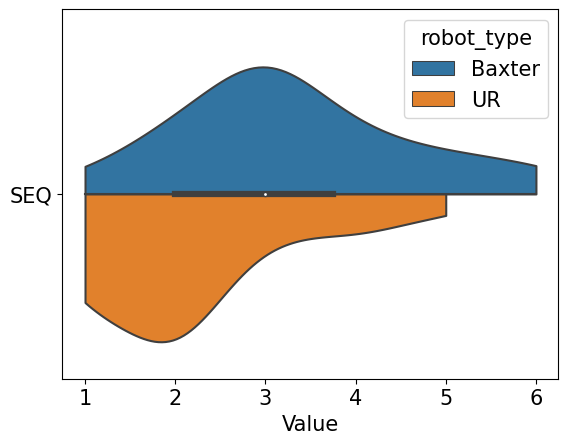}
\caption{Single Ease question violin plot for Baxter (orange) and UR3 (blue) (Higher number means easier to use)}
\label{fig:seq}
\end{figure}

Results of the Single Ease Question asked at the end of each task, in which a higher score means that TELESIM is easy to use, can be seen in Fig. \ref{fig:seq}. They show that the user was able to detect how well they performed and that their estimate is consistent with the result shown in Fig. \ref{fig:baxter_tower} and Fig. \ref{fig:ur_tower}. Specifically, Baxter obtained a mean of 3.32 with a standard deviation of 1.27, while UR3 obtained a mean of 2.19 with a standard deviation of 1.14. 
Furthermore, Fig. \ref{fig:seq} shows that no user gave the maximum score for UR3, while they did for Baxter. Additionally, the UR3 has a sharp decline in score after a SEQ score of 3, while Baxter's is more spread out.



\section{Conclusion and Future Works}

In this paper, we have investigated the performance of TELESIM by conducting a medium-scale user survey with 37 participants who were asked to build towers of 3 cubes by teleoperating robots. We tested TELESIM's modularity on two different robots with two different control modalities. Our experimental results show that TELESIM is modular, plug-and-play and user-friendly, as not only were we able to deploy it on 2 robots with different modalities, but most users were able to succeed by building at least once a tower of 3 cubes, with only 5 minutes of training, regardless of the control modality or robot used. We thus bridged the gap pointed out by Rea \& Seo \cite{rea_still_2022}, where they state that there is a lack of non-expert evaluation of robotic teleoperation for general tasks such as picking and placing common objects. TELESIM is available on GitHub\footnote{https://github.com/cvas-ug/telesim\_pnp}, allowing developers to perform teleoperation on their robots with minimal setup time.



Our underlying motivation for choosing direct teleoperation in this paper is to establish a baseline for further research on shared autonomy, which could combine human intuition and a high-level overview of a task while giving freedom to the robot to perform, for example, accurate picking and placing objects. 
Additionally, we plan to remove the constraint of having the VR headset behind the user and allow them to wear the headset to operate either in VR in the digital twin view of Isaac Sim or Augmented Reality by allowing the user to move around the robot and have different viewpoints while manipulating, thus enhancing the precision of the teleoperation.
However, the choice of control input is fundamental to success. Future work consists of carrying out a survey using the VR controller and the UR3 to dissociate the robot and control method; as for our current evaluation, we hypothesise they are closely linked.





\bibliographystyle{IEEEtran}
\bibliography{references, lib}

\end{document}